\pdfoutput=1

\documentclass[11pt]{article}

 \usepackage[]{emnlp2021}

\usepackage{times}
\usepackage{latexsym}

\usepackage[T1]{fontenc}

\usepackage[utf8]{inputenc}

\usepackage{microtype}
\usepackage{xr-hyper}
\usepackage{times}
\usepackage{latexsym}

\usepackage{xspace}
\newcommand{\sw}{\texttt{SWOW}\xspace}
\newcommand{\swow}{\texttt{SWOW}\xspace}
\newcommand{\cn}{\texttt{ConceptNet}\xspace}
\newcommand{\cnisw}{\texttt{CN$\cap$SW}\xspace}
\newcommand{\ck}{commonsense knowledge\xspace}
\hypersetup{final}

\usepackage{enumitem,amssymb,amsmath}
\usepackage{graphicx}
\usepackage{caption}
\usepackage{subcaption}

\usepackage{booktabs}
\usepackage{multirow}

\usepackage{comment}
\usepackage{siunitx}

\usepackage{csvsimple}
\usepackage{booktabs}
\usepackage{adjustbox}

\usepackage{tikz}
\usetikzlibrary{patterns,shapes.arrows}
\usepackage{pgfplots}
\pgfplotsset{compat=newest}
\usetikzlibrary{fit,shapes.geometric}
\usetikzlibrary{shapes,arrows,decorations.markings}
\usetikzlibrary{automata,positioning}
\usetikzlibrary{arrows.meta}
\usetikzlibrary{calc}
\usetikzlibrary{plotmarks}
\title{Commonsense Knowledge in Word Associations and ConceptNet}

\author{Chunhua Liu \quad Trevor Cohn \quad Lea Frermann \\School of Computing and Information Systems\\
        The University of Melbourne \\
        \texttt{chunhua@student.unimelb.edu.au}\\ \texttt{\{tcohn,lfrermann\}@unimelb.edu.au}\\
        }

\begin{document}
\maketitle
\begin{abstract}
Humans use countless basic, shared facts about the world to efficiently navigate in their environment. This {\it commonsense knowledge} is rarely communicated explicitly, however, understanding how commonsense knowledge is represented in different paradigms is important for both deeper understanding of human cognition and for augmenting {automatic} reasoning systems. This paper presents an in-depth comparison of two large-scale resources of general knowledge: \cn, an engineered relational database, and \swow a knowledge graph derived from crowd-sourced word associations. We examine the structure, overlap and differences between the two graphs, as well as the extent to which they encode situational commonsense knowledge. We finally show empirically that both resources improve downstream task performance on commonsense reasoning benchmarks over text-only baselines, suggesting that large-scale word association data, which have been obtained for several languages through crowd-sourcing, can be a valuable complement to curated knowledge graphs.\footnote{Code available at \url{ https://github.com/ChunhuaLiu596/CSWordAssociation}}
\end{abstract}

\section{Introduction}
Humans understand and navigate everyday situations with great efficiency using a wealth of shared, basic facts about their social and physical environment -- a resource often called commonsense knowledge \citep{liu_2004_conceptnet}. Advances in artificial intelligence in general, and natural language processing in particular, have led to a surge of interest in its nature: what constitutes commonsense knowledge? And despite this knowledge being rarely explicitly stated in text~\citep{Gordon2012ReportBias}, how can we equip machines with commonsense to enable more general inference~\citep{Davis2015CommonsenseRA}? Recently, large language models \citep{devlin2019bertpo,radford2019language} pre-trained on massive text corpora achieved promising results on commonsense reasoning benchmarks with fine-tuning, however, the lack of interpretability remains a problem. Augmenting them with commonsense knowledge can provide complementary knowledge \citep{Ilievski2021DimensionsOC,Safavi2021RelationalWK} and thus make models more robust and explainable \citep{lin2019kagnetkg}.

\tikzstyle{center}=[rectangle, draw,thick,
    text centered, rounded corners, minimum height=1em,draw, thick]
\tikzstyle{swow}=[rectangle, draw,thick,
    text centered, rounded corners, minimum height=0.8em,draw=cyan!50,fill=cyan!20]
\tikzstyle{cpnet}=[rectangle, draw,thick,
    text centered, rounded corners, minimum height=0.8em,draw=blue!50,fill=blue!20]
\tikzstyle{cpsw}=[rectangle, draw,thick,
    text centered, rounded corners, minimum height=0.8em,draw=green!50,fill=green!20]
    
\pgfplotsset{compat=newest}
\tikzstyle{annotation} = [rectangle, draw=none, fill=none, text centered, rounded corners]

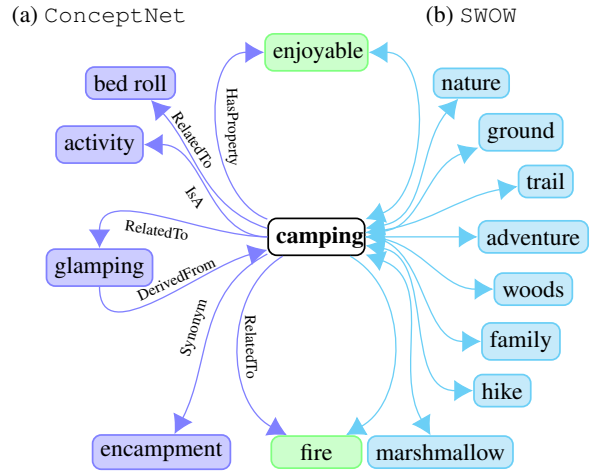
\begin{figure}[!t]

    \begin{adjustbox}{width=\columnwidth}
	\begin{tikzpicture}
    \node [center, text width=1.3cm, align=center] (camping) at (-0.5,0) {\textbf{camping}};
    \node [cpsw, above of= camping, node distance=3cm, xshift=0cm, text width=1.4cm, align=center] (enjoyable) {enjoyable};
        

    \node [cpnet, above of= camping, node distance=2.5cm, xshift=-3cm] (bed roll) {bed roll};
    \node [cpnet, left of= camping, node distance=3.5cm, yshift=1.5cm, align=center] (activity) {activity};
    \node [cpnet, left of= camping, node distance=3.5cm, yshift=-0.5cm, align=center] (glamping) {glamping};
    \node [cpnet, left of= camping, node distance=2.5cm, yshift=-3.5cm] (encampment) {encampment};
    \node [cpsw, below of= camping, node distance=3.5cm, yshift=0cm, align=center, text width=1.2cm] (fire) {fire};
    
    \path[-{Latex[black,length=5mm,width=2mm, angle=60:10pt,blue!50]},semithick, draw=blue!50, fill=blue!50, out=160, in=180 ] (camping)  edge  node[sloped, anchor=auto, above, scale=0.8,font=\small] {HasProperty}  (enjoyable);
    \path[-{Latex[black,length=5mm,width=2mm, angle=60:10pt,blue!50]},semithick, draw=blue!50, fill=blue!50, bend left, out=-60, in = -120] (camping)  edge  node[sloped, anchor=auto, above, scale=0.8,font=\small] {RelatedTo}  (fire);
     \path[-{Latex[black,length=5mm,width=2mm, angle=60:10pt,blue!50]},semithick, draw=blue!50, fill=blue!50, out=180, in=360] (camping)  edge  node[sloped, anchor=auto, below, scale=0.8,font=\small] {IsA}  (activity);
   \path[-{Latex[black,length=5mm,width=2mm, angle=60:10pt,blue!50]},semithick, draw=blue!50, fill=blue!50, bend right, out=30, in=180, xshift=-3cm] (camping)  edge  node[sloped, anchor=auto, above, scale=0.8, font=\small, xshift=-0.7cm] {RelatedTo} (bed roll);
    \path[-{Latex[black,length=5mm,width=2mm, angle=60:10pt,blue!50]},semithick, draw=blue!50, fill=blue!50, out=180, in =90] (camping)  edge  node[sloped, anchor=auto, below, scale=0.8, font=\small] {RelatedTo}  (glamping);
   
    \path[-{Latex[black,length=5mm,width=2mm, angle=60:10pt,blue!50]},semithick, draw=blue!50, fill=blue!50, out=270, in=195,] (glamping)  edge  node[sloped, anchor=auto, above, scale=0.8, font=\small,  xshift=0.5cm] {DerivedFrom}  (camping);
   
    \path[-{Latex[black,length=5mm,width=2mm, angle=60:10pt,blue!50]},semithick, draw=blue!50, fill=blue!50, bend left, out = -35, in=180 ] (camping)  edge  node[sloped, anchor=auto, above, scale=0.8, font=\small] {Synonym}  (encampment);

    \node[swow, above of = camping, node distance =2.5cm, xshift=2.5cm](nature) {nature};

    \node[swow, above of = camping, node distance=1.7cm, xshift=3.3cm](ground) {ground};
    \node[swow, right of = camping, node distance =3.5cm, yshift=0.0cm] (adventure) {adventure};
    \node [swow, right of= camping, node distance=3.5cm, yshift=-0.85cm] (woods) {woods};
    \node[swow, right of = camping, node distance =3.7cm, yshift=0.9cm] (trail) {trail};
     \node [swow, below of= camping, node distance=1.7cm, xshift=3.3cm, align=center] (family) {family};

    \node[swow, below of=camping, node distance = 3.5cm, xshift=2cm] (marshmallow) {marshmallow};
    \node[swow, below of = camping, node distance =2.5cm, xshift=3cm, align=left] (hike) {hike};

    \path[-{Latex[cyan!50,length=5mm,width=2mm, angle=60:10pt]},semithick,draw=cyan!50, fill=cyan!20, out=-30, in = 30] (camping) edge node [sloped, anchor=auto, above, xshift=0.5cm] {} (fire);
     \path[{Latex[cyan!50,length=5mm,width=2mm, angle=60:10pt]}-{Latex[cyan!50,length=5mm,width=2mm, angle=60:10pt]},semithick,draw=cyan!50, fill=cyan!20, out=20, in = 360] (camping) edge node [sloped, anchor=auto, above] {} (enjoyable);
     \path[{Latex[cyan!50,length=5mm,width=2mm, angle=60:10pt]}-{Latex[cyan!50,length=5mm,width=2mm, angle=60:10pt]},semithick,draw=cyan!50, fill=cyan!20, out=10, in = 230] (camping) edge node [sloped, anchor=auto, above] {} (nature);
     \path[{Latex[cyan!50,length=5mm,width=2mm, angle=60:10pt]}-{Latex[cyan!50,length=5mm,width=2mm, angle=60:10pt]},semithick,draw=cyan!50, fill=cyan!20, out=5, in = 200] (camping) edge node [sloped, anchor=auto, above] {} (ground);
      \path[-{Latex[cyan!50,length=5mm,width=2mm, angle=60:10pt]},semithick,draw=cyan!50, fill=cyan!20, out=5, in = 190] (camping) edge node [sloped, anchor=auto, below] {} (trail);
     \path[{Latex[cyan!50,length=5mm,width=2mm, angle=60:10pt]}-{Latex[cyan!50,length=5mm,width=2mm, angle=60:10pt]},semithick,draw=cyan!50, fill=cyan!20, out=360, in = 180] (camping) edge node [sloped, anchor=auto, above] {} (adventure);
     \path[{Latex[cyan!50,length=5mm,width=2mm, angle=60:10pt]}-{Latex[cyan!50,length=5mm,width=2mm, angle=60:10pt]},semithick,draw=cyan!50, fill=cyan!20, out=-20, in = 120] (camping) edge node [sloped, anchor=auto, below] {} (marshmallow);
    \path[-{Latex[cyan!50,length=5mm,width=2mm, angle=60:10pt]},semithick,draw=cyan!50, fill=cyan!20, out=360, in = 180] (camping) edge node [sloped, anchor=auto, above] {} (woods);
    
    \path[-{Latex[cyan!50,length=5mm,width=2mm, angle=60:10pt]},semithick,draw=cyan!50, fill=cyan!20, out=360, in = 180] (camping) edge node [sloped, anchor=auto, above] {} (family);
    
    \path[{Latex[cyan!50,length=5mm,width=2mm, angle=60:10pt]}-{Latex[cyan!50,length=5mm,width=2mm, angle=60:10pt]},semithick,draw=cyan!50, fill=cyan!20, out=-10, in =180] (camping) edge node [sloped, anchor=auto, above, xshift=0.5cm] {} (hike);

	\node[rectangle, draw=none, right of=enjoyable, anchor=base, yshift=0.5cm, xshift=1.5cm] (Conceptnet) {(b) \swow};
    \node[rectangle, draw=none, left of=enjoyable, anchor=base, yshift=0.5cm, xshift=-2.5cm] (Conceptnet) {(a) \cn};
	
	\end{tikzpicture}
    \end{adjustbox}
	\caption{\label{fig:camping_cn_sw} Sub-graphs centered around `\textbf{camping}' from (a)~\cn and (b)~\swow. Nodes in green are common to both KGs. Nodes on the left/blue (right/cyan) are unique to \cn (\swow). 
	}
\end{figure}

Attempts to capture general human knowledge include inventories of machine-readable logical rules~\cite{gordon2017formal} and large, curated databases which have been collected with the specific purpose to reflect either domain general (e.g., \cn; \citet{liu_2004_conceptnet}) or domain-specific (e.g., ATOMIC; \citet{sap2019atomicaa}) commonsense knowledge.

Word association norms are a third resource of explicit, basic human knowledge: in a typical study, human participants are presented with a {\it cue} word (e.g., `camping') and asked to produce one or more words that spontaneously come to mind (e.g., `hike', `nature', or `trail'). The resulting data sets of word associations have been used to explore the mental representations and connections underlying human language and reasoning \citep{Nelson2004TheUO,Fitzpatrick2006HabitsAR}, and have been shown to contain information beyond that present in text corpora \citep{de-deyne-etal-2016-predicting}. Word association studies have been scaled to thousands of cue words, tens of thousands of participants, and several languages, and hence provide a way of collecting diverse and unbiased representations. However, the extent to which they capture {\it commonsense} knowledge, and their utility for downstream applications have not yet been examined.

Prior work has probed the extent to which word associations contain lexical knowledge~\cite{de-deyne-etal-2016-predicting,Deyne2019TheW}, has systematically compared the relational coverage and overlap of curated knowledge bases~\cite{ilievski2020consolidating} (not including word associations data), and has investigated how much relational commonsense knowledge is present in language models~\cite{a-rodriguez-merlo-2020-word,vankrunkelsven2018predicting,da2019cracking}. We contribute to this line of research with an in-depth comparison of a dedicated commonsense knowledge base, and resources derived from spontaneous human-generated word associations. We systematically compare the most comprehensive, domain general, curated commonsense knowledge base (\cn) with the largest data set of English word associations (the ``Small World of Words''; \swow;~\citet{Deyne2019TheW}). We compare the two resources both in their structure and content, and apply them in downstream reasoning tasks.

Our contribution is important for three reasons. First, comparing explicitly engineered with spontaneously produced knowledge graphs can advance our fundamental understanding of the similarities and differences between the paradigms, and suggest ways to combine them. Second, recent progress in automatic commonsense reasoning largely focused on English and relies heavily on the availability of very large language models. These are, however, 
infeasible to train for all but a few high-resource languages. 
Finally, recent work has shown that the competitive performance of large language models on commonsense reasoning tasks is at least partially due to spurious correlations in language rather than genuine reasoning abilities; and that they perform close to random once evaluation controls for such confounds~\cite{elazar2021back}. Explicit representations of commonsense knowledge bases are thus have the potential to promote robust and inclusive natural language reasoning models.

In summary, our contributions are:

\begin{enumerate}
 \item We conduct an in-depth comparison of large-scale curated commonsense knowledge bases and word association networks, distilling a number of systematic differences which can inform future theoretical and empirical work.
 
 \item We analyze how much {\it commonsense} knowledge \cn and \sw encode, leveraging a human-created data set covering explicit situational knowledge. Our results suggest that \sw represents this knowledge more directly.
 
 \item We introduce \sw as a commonsense resource for NLP applications and show that it achieves comparable results with \cn across three commonsense question answering benchmarks.
\end{enumerate}
\section{Background}
\label{sec:graph_comparisons}
\smallskip

\subsection{Commonsense Knowledge Graphs}
Collecting and curating human commonsense knowledge has a rich history in artificial intelligence and related fields, motivated by the grand challenge of equipping AI systems with commonsense knowledge and reasoning ability~ \citep{Davis2015CommonsenseRA,lake_ullman_tenenbaum_gershman_2017}.  
Previous efforts have been put on collecting different aspects of commonsense knowledge, resulting in a variety of resources ranging from systems of logical rules \citep{Lenat1990BuildingLK}, over knowledge graphs \citep{liu_2004_conceptnet}, all the way to embedded representations \citep{Malaviya2019ExploitingSA}. Here, we focus on graph representations. Domain-specific examples include ATOMIC \citep{sap2019atomicaa} which focuses on social interactions in events, SenticNet \citep{Erik2020SenticNet6} which encodes sentiment-related commonsense knowledge. Recent work  also attempts to consolidate knowledge from multiple sources in order to improve knowledge coverage and utility~\citep{Ilievski2021CSKGTC, Hwang2020COMETATOMIC2O}. The largest domain-general \ck graph is \cn, which we will use in this study and describe in detail below.

\begin{table*}
    \centering
    \begin{tabular}[width=\textwidth]{lcccccc} 
        \toprule
    	\textbf{KG} & \textbf{\#Triples} & \textbf{\#Nodes} & \textbf{\#Relations} & \textbf{Density} & \textbf{Degree} & $H_N$\\
        \hline
    	\cn & 3,009,636 & 1,080,759 & 47 & \num{3.00e-6} & 2.78 & 23.28 \\
        \sw & 1,593,564 & 124,626 & 2 & \num{1.03e-4} & 12.78 & 18.07  \\
    	\bottomrule
    \end{tabular}
    \caption{\label{tab:statitics_cn_rel47_sw} Statistics of  \texttt{ConceptNet} and \texttt{SWOW} considered as directed graphs. Density is the graph density, Degree indicates the average node degree. $H_N$ indicates the node entropy. 
    }
\end{table*}


\paragraph{\cn}
All studies in this paper are based on the most recent \cn v5.6~\citep{Speer2017Conceptnet5.5}. \cn is a directed graph comprising over 3M nodes (aka concepts). Related concepts are connected with directed edges, which are labelled with one of 47 generic relation types. 
Figure~\ref{fig:camping_cn_sw}a shows a small subgraph, centered around the concept `camping'. Nodes are represented as free-text descriptions, which leads to a large node inventory and a sparsely connected graph. We filter out nodes that are not English, lowercase all descriptions and remove punctuation.  Row 1 in Table~\ref{tab:statitics_cn_rel47_sw} shows statistics of the resulting knowledge graph.


\subsection{Word Association Networks}
Human word associations have a long history in psychology and cognitive linguistics~ \citep{Deese1966TheSO,Deyne2019TheW}. Given a {\it cue} word, one or more spontaneous {\it responses} are elicited from study participants, shedding light on their mental associations. The resulting data sets, covering many cues and participants, can subsequently be turned into association {\it networks}, where each node corresponds to either a cue or response. Cues are connected to the responses they elicited through directed edges, which can be weighted (or filtered) by the number of participants who produced a particular response. The primary use-case of word associations has been to gain insights into the mental lexicon. A wealth of studies has shown the utility of word associations for predicting behavioural data including memory, lexical choice and semantic categorization~\cite{Nelson2000WhatIF,de2013associative,borge2010categorizing}, however, we are the first to inspect word association networks as a general commonsense knowledge base. Several association data sets have been collected, varying in coverage and target language (\newcite{Armstrong1973EAT,Nelson2004TheUO,jung2010network,Deyne2013BetterEO}). However, in order to consider this data as a general knowledge resource, it should (a)~cover a large set of diverse {cues}, and (b)~a large number of {responses} which are both diverse and reliable. Recently, the \emph{Small World of Words} project massively scaled word association collection through crowd-sourcing \cite{Deyne2019TheW}. In the remainder of this paper, we  use their English data set (``\sw'', described below). The \emph{Small World of Words} project has been scaled to 15 languages, suggesting its potential as a knowledge resource for NLP more generally.

\paragraph{Small World of Words}
\swow~\citep{Deyne2019TheW} is a word association network derived from a large collection of crowd-sourced cue-response pairs involving more than 90K participants and 12K cues. Given a cue, participants produced up to three responses. The resulting associations have been compiled into the \swow knowledge graph (Figure~\ref{fig:camping_cn_sw}b shows a small excerpt).
\swow edges are not labelled with relations. In some of our experiments, we adopt a basic set of two relations using the associative directions, namely {\it forward associations} from a cue to a response, and {\it mutual associations}  for pairs where the reverse is also included in \sw. We use the official, pre-processed release of \sw.\footnote{\url{https://smallworldofwords.org/en/project/research}} We remove ``NA'' responses, lowercase all node descriptions and remove punctuation. Row 2 in Table~\ref{tab:statitics_cn_rel47_sw} shows statistics of the resulting graph.

In the remainder of this paper, we conduct three experiments to compare \sw and \cn from different dimensions, ranging from the intrinsic graph properties (\S\ref{sec:experiment_one}), to the coverage of encoded commonsense knowledge (\S\ref{sec:experiment_two}), and their utility for  downstream commonsense reasoning tasks (\S\ref{sec:experiment_three}).

\section{Experiment 1: Intrinsic Comparisons}
\label{sec:experiment_one}
A knowledge graph  consists of a set of nodes $\mathcal{N}$ and edges $\mathcal{E}$, comprising triples $\langle e, r, e'\rangle$ to denote a directed edge from head node $e$ to tail node $e'$ labelled with relation $r \in \mathcal{R}$. We denote $\mathcal{E}(e)$ as the incoming and outgoing edge set for node $e$, $\mathcal{E}(r)$ as the set of edges with relation~$r$, and $|\cdot|$ as the size of a set. Here, we consider the specific knowledge graphs \cn and \sw, and begin by comparing the intrinsic properties: their typology and content encoded.


\subsection{Knowledge Graph Structure}
\label{ssec:knowleddge_graph_structure}
\cn is a substantially larger graph than \sw, with about eight times as many nodes and 1.9$\times$ as many edges (cf.,~Table \ref{tab:statitics_cn_rel47_sw}). We compare sparsity in terms of (1)~{\it graph density}, 
\begin{equation*}
    \frac{|\mathcal{E}|}{|\mathcal{N}|(|\mathcal{N}|-1)},
\end{equation*}
and (2)~{\it node degree} as the average total of incoming and outgoing edges~\citep{Malaviya2019ExploitingSA}. Table~\ref{tab:statitics_cn_rel47_sw} shows that \sw has 39$\times$ the density and 4$\times$ the average node degree of \cn: Even though \sw is smaller than \cn, it is substantially more densely connected.

\paragraph{Node Distribution.}
To better understand the distribution of nodes in the KGs, we  measure node diversity via the entropy of the node distribution~($H_N$; \newcite{pujara-etal-2017-sparsity}):
\begin{equation*}
\begin{aligned}
    H_N &=\sum_{e \in N}-P(e) \log P(e),\\
\end{aligned}
\end{equation*}
where $P(e) = |\mathcal{E}(e)| / |\mathcal{E}| $ is the fraction of edges incident on the node.

Higher $H_N$ indicates a more uniform node distribution where many nodes are connected, whereas lower entropy suggests a skewed distribution, where few nodes are highly connected. Table~\ref{tab:statitics_cn_rel47_sw} shows a lower $H_N$ for \sw, i.e., nodes are less uniformly connected. This is because \sw is by construction more structured than \cn: its  12K \textit{cue} nodes are densely connected to a much larger number of (sparsely connected) \textit{response} nodes. 


\paragraph{Node and Edge Overlap.}
\label{ssec:shared_knowledge}

How large is the overlap between \cn and \swow? We quantify the overlap of individual nodes in the two KGs, based on exact string match.\footnote{
String matching is arguably a simplistic way of matching concepts across two KGs. Advanced methods of concept resolution could leverage embedding methods. We leave this interesting direction for future work.} 
We find 58\%  (71K) of the nodes in the smaller \sw are present in \cn (conversely, 7\% of the nodes in \cn are present in \swow). Over 40\% of the concepts in \swow are not present in \cn, which is perhaps expected given their very distinct methods of construction, but motivates further in-depth comparison (Section~\ref{subsec:edge_type_comparison}). Moving on to edge overlap, which we measure over undirected head-tail pairs,%
\footnote{Edge comparison ignores direction as many relations can naturally be inverted, e.g., `part of' and `has part'. Consequently linking concepts in either direction is considered to be correct.}
we find that 6\% of edges in \cn are present in \sw, and 15\% of the edges in \sw are present in \cn. 
This low overlap demonstrates that human associations indeed elicit connections among words missed in the large database \cn. 
For the 71K overlapping nodes, we further find that 691K connections exist in \cn, and 1.5M connections in \sw, covering 95\% of all \sw edges. This again suggests that \sw is more comprehensive than \cn.

\pgfplotstableread[row sep=\\,col sep=&]{
    rank & tags & CN & SW & CN_Ratio & SW_Ratio & Delta \\
    1 & NOUN & 523580 & 49077 & 0.4844 & 0.3937 & -9.07 \\
	2 & NP & 186841 & 28321 & 0.1729 & 0.2272 & 5.43 \\
	3 & VERB & 122486 & 14841 & 0.1133 & 0.1191 & 0.58 \\
	4 & VP & 86035 & 8801 & 0.0796 & 0.0706 & -0.9 \\
	5 & ADJ & 80485 & 8784 & 0.0745 & 0.0705 & -0.4 \\
	6 & ADV & 22008 & 1870 & 0.0204 & 0.015 & -0.54 \\
	7 & INTJ & 15151 & 1612 & 0.014 & 0.0129 & -0.11 \\
	8 & S & 8977 & 1520 & 0.0083 & 0.0122 & 0.39 \\
	9 & PP & 7118 & 3000 & 0.0066 & 0.0241 & 1.75 \\
	10 & ADJP & 6978 & 2374 & 0.0065 & 0.019 & 1.25 \\
}\DataCNSW

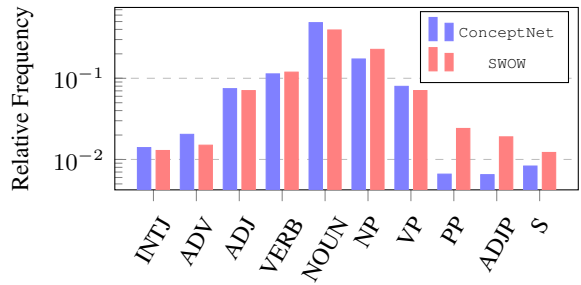
\begin{figure}[t!]
    \pgfplotsset{
        tick label style={font=\footnotesize},
        label style={font=\footnotesize},
        legend style={font=\tiny},
        }
\begin{center}
    
\begin{tikzpicture}[baseline]
    \begin{semilogyaxis}[
            name = plot1,
            ybar,
            ymin=0,
            bar width=5pt,
            width=0.48*\textwidth,
            height=4cm,
            xtick pos = left, 
            ytick pos = left, 
            log origin=infty,  
            symbolic x
            coords={INTJ,ADV,ADJ,VERB,NOUN,NP,VP,PP,ADJP,S},
            xticklabel style={rotate=60},
            xtick=data,
            ylabel={Relative Frequency},
            ymajorgrids=true, 
            grid style=dashed,
            label style={font=\footnotesize},
        ]
   
        \addplot[color=blue!50!white,fill=blue!50!white] table[x=tags,y= CN_Ratio]{\DataCNSW};
        \addplot[color=red!50!white,fill=red!50!white] table[x=tags,y=SW_Ratio]{\DataCNSW};
        \legend{\texttt{\cn}, \texttt{\sw}};
    \end{semilogyaxis}
    \end{tikzpicture}
\caption{\label{fig:tag_node_pos_phrase_cn_sw} \normalfont The distribution of syntactic tags on \texttt{ConceptNet} and \texttt{SWOW} for the 10 most frequent tags.}
\end{center}
\end{figure}
\subsection{Knowledge Graph Content}
\label{subsec:edge_type_comparison}
Having established the structural characteristics of \cn and \sw, we will now focus on their respective encoded knowledge.

\subsubsection{Conceptual Content} 

Nodes in \cn and \sw express concepts as words or short phrases. We compare: (1) the distributions of the syntactic categories for concepts over the two knowledge bases; (2) the occurrences of concepts in two KGs in large corpus. 

We use a constituency parser to predict the syntactic phrase or part of speech (POS) tag for a concept string.\footnote{We use the parser of~\newcite{Kitaev-2018-SelfAttentive} as implemented in Spacy.} 
The relative prevalence of the 10 most frequent syntactic types is shown in Figure~\ref{fig:tag_node_pos_phrase_cn_sw}. While the overall distribution is similar in both KGs, two patterns emerge. First, even though both KGs are dominated by nominal nodes, \sw's distribution over POS types is less skewed, suggesting concepts are more diverse. Secondly, the proportion of phrasal concepts compared to single-word concepts tends to be higher in \swow compared to \cn.

Next we examine the corpus frequency of concepts in \cn and \sw using the Google n-gram corpus.\footnote{\url{https://books.google.com/ngrams}} 
Many \cn concepts were not present in this large corpus (24\%), versus 1.5\% for \swow.
Of those concepts that could be found, concepts in \swow are on average $7\times$ more common than those in \cn.
We conclude that \sw concepts are generally common, while \cn includes more obscure concepts. 


\subsubsection{Relational Content}
We cannot directly compare relation distributions between \cn and \sw because \sw does not have labelled relations. Instead, we inspect the intersection of the two graphs (\cnisw) as all head-tail pairs that are shared between the graphs ($|\mathcal{E}_{\text{\cnisw}}|$=190K), labelled with their \cn relations. We use \cnisw as a proxy of the relations in \sw. 
For each relation type $r$, we compare its relative frequency in the full \cn ($f^r_{\texttt{CN}}$) against its recall in \cnisw ($\text{recall}^r$), where
\begin{equation*}
 \begin{aligned}
    f^r_{\text{\texttt{CN}}} &= \frac{|\mathcal{E}_{\texttt{CN}}(r)|}{|\mathcal{E}_{\texttt{CN}}|}, & 
    \text{recall}^r &= \frac{|\mathcal{E}_{\cnisw}(r)|}{|\mathcal{E}_{\texttt{CN}}(r)|}.
 \end{aligned}
\end{equation*}

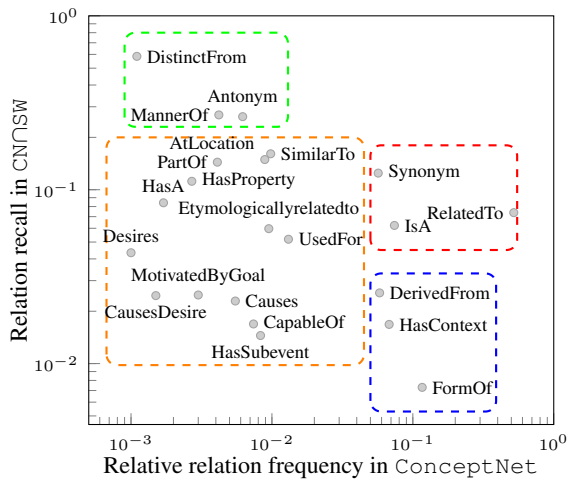
\begin{figure}[!t]
  \centering
\begin{tikzpicture}
title={\texttt{patch type=quadratic spline}}]
\begin{axis}[
    xlabel={Relative relation frequency in \cn},
    ylabel={Relation recall in \texttt{CN$\cap$SW}},
    label style={font=\small},
    xlabel style={
      yshift=1ex,
      name=label,
      },
    ylabel style={ yshift=-1ex, name=label,}, 
    tick label style={font=\tiny},
    xtick pos = left, 
    ytick pos = left, 
    xmode=log,
    ymode=log,
    ymax=1,
    xmax=1,
    width=\columnwidth,
    height=7cm,
    colormap={summap}{
                color=(black!20); color=(black!20); 
                color=(black!20); color=(black!20) 
                color=(black!20) color=(black!20)
                color=(black!20)
                },
    colorbar style={
            at={(0.1,1.0)},               
            anchor=below south west,    
            width=0.8*\pgfkeysvalueof{/pgfplots/parent axis width},
              xmax=0.61,
             xticklabel=$10^{\pgfmathparse{\tick}\pgfmathprintnumber\pgfmathresult}$, 
        },
     colorbar/width=1.5mm,
    ]

  \addplot+[
    only marks,
    color=black, 
    mark=*,
    mark size=1.5pt,
    nodes near coords*={\Label},
    nodes near coords style={font=\scriptsize, anchor=south},
    visualization depends on={value \thisrow{rel} \as \Label},
    ] table [x=CN_Percentage, y=Overlap_recall, col sep=comma,] {data1.csv}; \label{plot_one}
    
    \addplot+[
    only marks,
    color=black, 
    mark=*,
    mark size=1.5pt,
    nodes near coords*={\Label},
    nodes near coords style={font=\scriptsize, anchor=east},
    visualization depends on={value \thisrow{rel} \as \Label},
    ] table [x=CN_Percentage, y=Overlap_recall, col sep=comma,] {data2.csv}; \label{plot_one_one}

  \addplot+[
      only marks,
      color=black, 
      mark=*,
      mark size=1.5pt,
      nodes near coords*={\Label},
      nodes near coords style={font=\scriptsize, anchor=north},
      visualization depends on={value \thisrow{rel} \as \Label},
      ] table [x=CN_Percentage, y=Overlap_recall, col sep=comma,] {data3.csv}; \label{plot_two}
    \addplot+[
        only marks,
        color=black, 
        mark=*,
        mark size=1.5pt,
        nodes near coords*={\Label},
        nodes near coords style={font=\scriptsize, anchor=west},
        visualization depends on={value \thisrow{rel} \as \Label},
        ] table [x=CN_Percentage, y=Overlap_recall, col sep=comma,] {data4.csv}; \label{plot_three}

        \addplot+[
        only marks,
        color=black, 
        mark=*,
        mark size=1.5pt,
        nodes near coords*={\Label},
        nodes near coords style={font=\scriptsize, anchor= west},
        visualization depends on={value \thisrow{rel} \as \Label},
        ] table [x=CN_Percentage, y=Overlap_recall, col sep=comma,] {data5.csv}; \label{plot_three_one}

      \addplot+[
          only marks,
          color=black, 
          mark=*,
          mark size=1.5pt,
          nodes near coords*={\Label},
          nodes near coords style={font=\scriptsize, anchor=south east},
          visualization depends on={value \thisrow{rel} \as \Label},
          ] table [x=CN_Percentage, y=Overlap_recall, col sep=comma,] {data6.csv}; \label{plot_four}

    \draw [rounded corners, green,thick,dashed,] (0.0009,0.23) rectangle(0.013,0.8); %
    \draw [rounded corners, red,thick,dashed,] (0.05,0.045) rectangle(0.55,0.18); %
    \draw [rounded corners, blue,thick,dashed,] (0.05,0.0053) rectangle(0.39,0.033);
    \draw [rounded corners, orange,thick,dashed,] (0.00067,0.0098) rectangle(0.045,0.20);%
\end{axis}

\end{tikzpicture}
\caption{\label{fig:relation_distribution} The correlation between the relative frequency of \cn relations and their recall in the overlap subgraph, \texttt{CN$\cap$SW}.
}
\end{figure}
Figure~\ref{fig:relation_distribution} plots the correlation between the $f^r_{\texttt{CN}}$, and their recall in \cnisw. First, we observe a discenerable correlation between the majority of low- to medium frequency relations in \cn and their recall in \cnisw (\textcolor{orange}{orange box} on bottom left).
These relations cover largely semantic associations pertaining to the appearance, use or situational contexts of concepts. Second, none of the six most frequent relation types in \cn (right part of the plot), are highly recalled in \cnisw. Out of these, relations indicating (near) synonymy retain a medium recall (\textcolor{red}{red box} on middle right), while morphosyntactic relations are less prevalent (\textcolor{blue}{blue box} on bottom right). These relations are prevalent in \cn, because it is derived to a large part from structured, linguistic resources like Wiktionary.\footnote{74\% of \cn edges origin from Wiktionary.} Word associations, on the other hand, are known to be dominated by semantic relations~\cite{mollin2009combining}.
Third, the relations with highest recall in \cnisw tend to be infrequent in \cn (\textcolor{green}{green box} on top left). 
Two of these relations focus on {\it differences} indicating that humans associate contrasting concepts (such as `hot'$\rightarrow$`not cold'; \citet{Deese1964TheAS,Herbert1970WordAsso}).
We also found that the proportion of negated edges in \sw (0.7\%; N=11K) is more than twice the proportion in \cn (0.3\%; N=11.5K), and that only 4\% of them overlap.\footnote{Negated nodes were identified based on a list of negation markers, cf.,~Appendix~\ref{sec:negation_list}.} Representations of antonyms and negations have traditionally been difficult to infer from text, suggesting that \sw could be a valuable complementary knowledge source to fill this gap.


\input{figures/narrative_kg_graphs_183_tikz.tex}

\section{Experiment 2: Coverage of Commonsense Knowledge}
\label{sec:experiment_two}
This section probes \cn and \sw specifically for (situational) {{\it commonsense} knowledge}. Daily activities such as {\it doing the laundry} or {\it visiting the doctor} involve a wealth of general knowledge touching on causal, temporal, physical, or social knowledge which is rarely explicitly stated~\cite{Mostafazadeh2016ACA,rashkin-etal-2018-modeling,ostermann-etal-2018-mcscript,ostermann-etal-2019-mcscript2}. 
We leverage the MCScript2.0 data set~\citep{ostermann-etal-2019-mcscript2}, a large collection of {\it explicit} descriptions of everyday scenarios, and investigate whether \cn and \sw encode the commonsense knowledge underlying these situations. In particular, we test~whether the knowledge graph structure underlying MCScript scenarios is retained in \cn and \sw. Figure~\ref{fig:narrative_kg_graphs_183} shows an example of a scenario (a), derived graph representation (b), and a subset of corresponding associations in \cn (c) and \sw (d).


\subsection{Method}
\label{subsec:narrative_commonsense_knowledge_comparison}

\pgfplotsset{compat=1.12}
\begin{figure}[t!]
    \pgfplotsset{
        tick label style={font=\small},
        label style={font=\small},
        }
\begin{center}
\begin{tikzpicture}
        \pgfplotstableread{figures/shortest_path_length_cn.dat}\data
        \pgfplotstablegetrowsof{\data}
        \pgfmathsetmacro{\N}{\pgfplotsretval}
        \begin{axis}[
            ybar,
            ymin=0, 
            width=\columnwidth,
            height=4.5cm,
            xtick pos = left, 
            ytick pos = left, 
            yticklabel={%
                \pgfmathparse{\tick*100}%
                \pgfmathprintnumber{\pgfmathresult}\,},
            xlabel = {Shortest Path Length},
            ylabel = {Percentage(\%)}, 
            ylabel near ticks,
            yticklabel style = {font=\tiny},
            xticklabel style = {font=\tiny},
            ylabel style={font=\small},
            xlabel style={font=\small},
            legend style={font=\tiny}
        ]
            \addplot+[hist={data=x, bins=50}, 
                     y filter/.expression={y/\N},
                     color=blue!50!white,fill=blue!50!white,
                     opacity=0.7,
                     draw opacity=0
                      ] file {figures/shortest_path_length_cn.dat};
            \addplot+[hist={data=x, bins=50}, 
                     y filter/.expression={y/\N},
                     color=red!50!white,fill=red!50!white,
                     opacity=0.7,
                     draw opacity=0
                      ] file {figures/shortest_path_length_sw.dat};
            \legend{\cn, \sw};
        \end{axis}
    \end{tikzpicture}
\end{center}

\caption{ \label{fig:shortest_path} The length distribution of average shortest paths in \cn and \sw for edges from MCScript graphs.}
\end{figure}
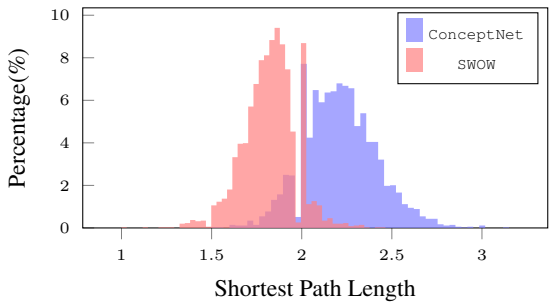

\paragraph{MCScript2.0} is a collection of 3,487 short narrative descriptions covering 200 every-day scenarios of varying complexity (e.g., {\it cleaning the floor} vs {\it growing vegetables})~\cite{ostermann-etal-2019-mcscript2}. The descriptions were crowd-sourced, and authors were instructed to describe the underlying scenarios ``as if talking to a child''~\cite{ostermann-etal-2018-mcscript}. Thus by design MCScript narratives spell out commonsense knowledge more explicitly than most text corpora. 
Following prior work on modelling narrative scripts, we posit that narrative chains are fundamentally characterized in terms of their events and participants~\cite{chambers2009unsupervised,frermann2014hierarchical}. 
We recover this information using semantic role labelling (SRL), and identify predicates, ARG0s and ARG1s in each narrative. We use the resulting set of spans and their relations to transform each narrative (Figure~\ref{fig:narrative_kg_graphs_183}a) into a {\it script graph} (Figure~\ref{fig:narrative_kg_graphs_183}b). Spans constitute the nodes of the script graph,\footnote{Pronouns and stopwords are excluded. We use the SRL model of~\citet{Shi2019SimpleBM} implemented in AllenNLP.} and edges correspond to (undirected) SRL dependencies.

\paragraph{Graph mapping} 
We assume that predicates-argument relations in detailed descriptions of everyday events (e.g., `water' and `garden') are instances of common sense knowledge, and should hence be directly encoded in a KG. 
We posit that the shorter the paths, the more directly the necessary scenario-specific commonsense knowledge is encoded in the target KG. E.g.,  we find a direct 1-hop connection between `water' and `garden' in \sw (Figure~\ref{fig:narrative_kg_graphs_183}d), while the path between the concepts in \cn is less direct (2-hops;  Figure~\ref{fig:narrative_kg_graphs_183}c). 

More technically, we project the MCScript graphs onto \sw and \cn as follows. For all directly connected node pairs in a MCScript graph, we identify their corresponding concepts in \sw and \cn, respectively, by exact string match. In our current study we are primarily interested in {\it whether} (not how) node pairs from the MCScript graph exist in the target KGs. {We therefore treat all graphs as undirected, and retrieve the shortest path between the two nodes in the target KGs.} 
Figure~\ref{fig:narrative_kg_graphs_183}c and \ref{fig:narrative_kg_graphs_183}d show a subset of the shortest paths retrieved from \cn and \swow, respectively, for the MCScript graph in \ref{fig:narrative_kg_graphs_183}b (bolded nodes). See Appendix ~\ref{sec:full_list_of_shortest_paths} for the full list of retrieved paths from \cn and \sw.

\subsection{Results}
Figure~\ref{fig:shortest_path} presents the distribution over shortest path lengths, averaged over the full MCScript data set. We observe that paths, on average, are shorter in \sw compared to \cn, suggesting that the situational commonsense associations in MCScript are more directly encoded in \sw. The example shortest paths shown in Figure~\ref{fig:narrative_kg_graphs_183}c (\cn) and d (\swow) further illustrate the associations in the two commonsense resources. The associations imposed in paths of length ${>}1$ are meaningful across the board, but differ across the KGs: for example, the connection between `bee' and `arrive' is further elaborated in \cn by explaining that in order to arrive, the bee needs to leave (from a plausible location `branch'); \swow on the other hand imputes information on the mode of travel (`flying'). 

As a first exploration into evaluating the extent of commonsense knowledge in commonsense KGs, our method has several weaknesses to be addressed in future work: the string mapping from text to \ck could be replaced with more flexible, embedding-based methods; the script graphs themselves could be improved and enriched with more semantic roles, or higher-level narrative structure as captured for instance in Rhetorical Structure Theory~\cite{taboada2006rhetorical}. Similarly, the mapped graphs could incorporate edge directions and/or labels. Ultimately, the mapped paths (most interestingly those mapped to paths of length $>1$) will need to be validated and interpreted by human annotators. We believe that leveraging explicit human-created commonsense data sets, like MCScript2.0, opens interesting avenues to understand the commonsense knowledge present in word associations.

\section{Experiment 3: Commonsense QA}
\label{sec:experiment_three}
In this section, we explore the utility of commonsense knowledge in \cn and \sw in commonsense question answering (CQA) tasks. We incorporate the two KGs into representative and competitive CQA models from the recent literature \citep{Wang2020ConnectingTD,Feng2020ScalableMR}, and apply them to three benchmark data sets. We emphasize that the goal of this study is not competing on leaderboards. Current state-of-the-art models leverage very large language models with billions of parameters \citep{Khashabi2020UnifiedQACF}, and often draw on additional external resource such as Wiktionary \citep{Xu2020FusingCI}.
Instead, we explore the utility of \sw and \cn in a selection of representative moderately complex models.

\subsection{Experimental setup}
\label{sec:models}
\begin{table*}[!ht]
    \centering
    \small
    \begin{tabular}{lp{10cm}l}
	\toprule
	 Dataset       & Example & Train / Dev / Test split \\\midrule
     CSQA & What do all humans want to experience in their own home?\newline
(a) {\bf feel comfortable}, (b) work hard, (c) fall in love, (d) lay eggs, (e) live forever& 8,500/1,221/1,241 \\
     OBQA   &What is a source of energy?\newline (a) bricks, (b) \textbf{grease}, (c) cars, (d) dirt  & 4,957/500/500 \\
     MCScript2.0   &When did small plants grow?\newline (a) two days, (b) \textbf{after seeds were planted} &  14,191/2,020/3,610\\

	\bottomrule
    \end{tabular}
    \caption{Details on the benchmarks CSQA, OpenbookQA and MCScript2.0: One example QA-pair per dataset (correct answer in boldface) and sizes of the respective train/dev/test splits. The paragraph of MCScript2.0 example is shown in Figure~\ref{fig:narrative_kg_graphs_183}.
    }
    \label{tab:dataset_examples}
\end{table*} 
\begin{table*}[!t]
    \small
    \centering
    \begin{adjustbox}{width=\textwidth}
    \begin{tabular}{lcccccc}
	\toprule
	Models & \multicolumn{2}{c}{CSQA}  &\multicolumn{2}{c}{ OBQA}  &\multicolumn{2}{c}{MCScript2.0}  \\
	 & \cn & \sw & \cn & \sw & \cn & \sw \\ \cmidrule(lr){2-3} \cmidrule(lr){4-5} \cmidrule{6-7}
	ALBERT & \multicolumn{2}{c}{73.78 (\(\pm \)0.79) }  & \multicolumn{2}{c}{63.47 (\(\pm \)1.42) } &  \multicolumn{2}{c}{93.62 (\(\pm \)0.44) }  \\ \cmidrule(lr){2-3} \cmidrule(lr){4-5} \cmidrule{6-7} 
	+ GconAttn & 74.03 (\(\pm \)0.46) & 74.05 (\(\pm \)0.50) &  65.13 (\(\pm \)2.16) & 65.87 (\(\pm \)1.21) & {\bf 93.91} (\(\pm \)0.50) & 93.84 (\(\pm \)0.35)\\
	+ RN & {\bf 75.64} (\(\pm \)0.70) & 74.40 (\(\pm \)0.37) &  64.73 (\(\pm \)2.10) & {\bf 66.40} (\(\pm \)1.00) & 93.53 (\(\pm \)0.11) & 93.49 (\(\pm \)0.22)  \\
	\bottomrule
\end{tabular}
\end{adjustbox}
\caption{\label{tab:results_main_table_csqa} Test accuracy on CSQA, OBQA and MCScript2.0. We report performance of ALBERT, and augment it with two KG-aware models using either \cn or \swow. Results are averages of the best three out of six runs (based on dev set performance); standard deviations reported in brackets.
}
\end{table*}
\paragraph{Datasets}
We consider three standard multiple-choice CQA benchmark datasets. \textbf{CommonsenseQA} (CSQA; \newcite{talmor-etal-2019-commonsenseqa}) contains commonsense questions generated by crowd workers on the basis of sub-graphs in \cn, giving \cn an inherent advantage over \swow. The QA-pairs in this dataset require various commonsense skills, and distractor answers were carefully selected to share semantic associations with the key concepts in a question. 
\textbf{OpenBookQA} (OBQA; \newcite{Mihaylov2018CanAS}) consists of question-answer pairs along with paragraphs from elementary-level science books. Following previous work~\citep{Wang2020ConnectingTD,Feng2020ScalableMR}, we disregard the paragraphs, and apply our models to question-answer pairs directly. 
Building on our analysis in Section~\ref{sec:experiment_two}, we also apply our models to the \textbf{MCScript2.0} QA benchmark~\citep{ostermann-etal-2019-mcscript2}. Each task consists of a story, and a question paired with two answer options. 
We use the in-house data split by \newcite{lin2019kagnetkg} for CSQA and the official splits for the other data sets. Table~\ref{tab:dataset_examples} presents data set statistics, as well as an example from each dataset.

\paragraph{Models}
A typical QA system consists of three modules: (1) a KG encoder which maps a subgraph spanning the concepts in the question/answer to a fixed-dimensional knowledge embedding 
$\mathbf{k}$;(2) a text encoder, which maps question $q$ and answer $a$ to a fixed-dimensional language embedding $\mathbf{c}$; and (3) a scoring module which scores each answer option given $\mathbf{c}$ and $\mathbf{k}$, and returns the highest scoring answer as a prediction. 

We experiment with two KG encoders: {\sf GconAttn}~\citep{Wang2019ImprovingNL} maps question and answer concepts to pre-trained concept embeddings, and then aligns them using concept-level attention and pooling. GconAttn is a relation-free model, leveraging only mentioned concepts from a KG. To disentangle the impact of relation labels, we also experiment with {\sf Relation Networks (RN)}~\citep{Santoro2017ASN}, which embed concepts using context-aware path-level attention over path embeddings. Each path embedding  encodes the path between some question concept and answer concept. We use one-hop and two-hop paths in our experiments. We refer the reader to the respective papers for full model details, as well as  Appendix~\ref{sec:re_production_results} which details our full parameter settings and reproduction results.
Both KG encoders require {\em node embeddings}. We use RoBERTa-Large \citep{Liu2019RoBERTaAR} to obtain a node embedding matrix, separately for \cn and \swow. 
Specifically, for each node $\mathcal{E}_i \in \mathcal{E}$, we feed the sequence of \text{[\texttt{CLS}]  $ + \, \mathcal{E}_i +$ [\texttt{SEP}]} to RoBERTa and use the last layer representation of [\texttt{CLS}] as its embedding. 

RN also requires {\em relation embeddings}. For each of our KGs (\cn, \swow), we obtain a separate relation embedding matrix $\mathbf{R}$ with {\sf TransE}.\footnote{Using OpenKE \url{https://github.com/thunlp/OpenKE}. We do not use TransE node embeddings, as they were outperformed by RoBERTa embeddings in preliminary tests.}
Following previous work on CQA~\citep{lin2019kagnetkg,Wang2020ConnectingTD}, we select 31 out of \cn's 47 relation types which proved helpful for CQA, and merge the remaining relations into 17 types. We use forward- and mutual associations for \swow.  Following previous work ~\citep{Malaviya2019ExploitingSA,Wang2020ConnectingTD}, we densify both \swow and \cn by adding for each relation $\mathcal{R}_i$ an additional relation type indicating its reverse $\mathcal{R}_i^{-1}$. For example, a relation $\mathcal{R}_i{=}\text{part\_of}$ would be complemented with $\mathcal{R}_i^{-1}{=}\text{has\_part}$. 

We fix the text encoder to {\sf ALBERT-xxlarge-v2}~\citep{lan2020albertal}, which performed competitively in recent work~\citep{Wang2020ConnectingTD}. 
All models are run six times and we report results using the best three models, as judged by training loss; this method is used to remove outliers resulting from instability of training.
All results are our own re-runs using the official implementations from \url{https://github.com/INK-USC/MHGRN}, and are largely comparable with those reported in the literature. See Appendices~\ref{sec:hyper_parameters},~\ref{sec:re_production_results} for further details.

\subsection{Results}
\label{sec:experimental_results}

Experimental results in Table~\ref{tab:results_main_table_csqa} show that both knowledge-augmented models outperform the language baseline for all data sets except RN on MCScript2.0. The path-aware RN achieves the best performance for both CSQA and OBQA. All models (including the text-only baseline) show comparative and very high performance on MCScript2.0, suggesting that it is a simpler task compared to the other two. Our models outperform the current state-of-the art on MCScript2.0 by up to 3.24\% (absolute).\footnote{\url{https://coinnlp.github.io/task1.html}} More importantly, models incorporating either of \cn or \swow achieve similar performance across the board. Recall that CSQA is derived from \cn edges, putting \sw at a disadvantage. \sw performs best on OBQA which is independent of both KGs. {We measure the significance of differences in performance between the text-only baseline and the RN models on CSQA and OBQA using Student's t-test. We find that the RN models outperform the text-only model significantly ($p{<}0.05$) with both \cn and \swow as underlying KG.\footnote{The only exception is ALBERT vs RN with \cn on OBQA where~$p{=}0.07$.} There is no significant difference between the \cn-based and the \swow-based RN models ($p{>>}0.05$).}
These results provide initial evidence that \sw can be a valuable alternative source of commonsense knowledge to \cn for downstream NLP tasks. 

\pgfplotstableread[row sep=\\,col sep=&]{
	NR & RN-mean & RN-variance & PG-Full-mean & PG-Full-variance \\
	17 & 75.64 & 0.7 & 76.85 & 0.61 \\
	7 & 76.15 & 0.34 & 77.32 & 0.16 \\
	1 & 75.85 & 0.72 & 75.74 & 0.98 \\
	0 & 74.83 & 1.1 & 74.11 & 0.49 \\
}\CNCSQAIHtest

\pgfplotstableread[row sep=\\,col sep=&]{
	NR & RN-mean & RN-variance & PG-Full-mean & PG-Full-variance \\
    17 & 64.73 & 2.1 & 67.67 & 1.68 \\
	7 & 65.2 & 1.4 & 66.53 & 2.05 \\
	1 & 66.4 & 1.04 & 67.53 & 1.3 \\
	0 & 65.3 & 0.99 & 65.8 & 1.04 \\
}\DataCNOBQAtest

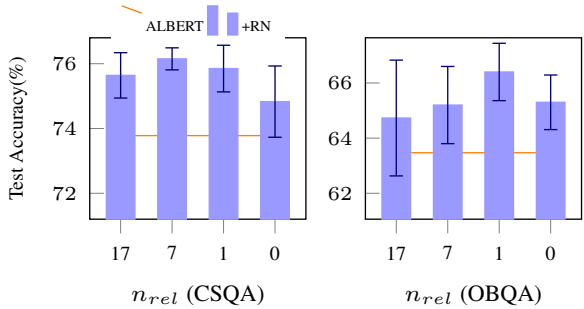
\begin{figure}[!tb]
\begin{adjustbox}{width=\columnwidth}
\begin{tikzpicture}[baseline]
    \begin{axis}[
            name = plot1,
            ybar,
            bar width=.3cm,
            width=0.5*\columnwidth,
            height=3.5cm,
            enlargelimits=0.2, 
            legend style={at={(0.5,1.22)},anchor=north, font=\tiny},
            xtick pos = left, 
            ytick pos = left, 
            symbolic x coords={17,7,1,0},
            xtick=data,
            y tick label style={
                        /pgf/number format/set thousands separator={}%
                        },
            ymin=72,
            ymax=76,
            ylabel={Test Accuracy(\%)},
            ylabel style = {font=\tiny },
            xlabel style = {font=\scriptsize},
            yticklabel style = {font=\tiny},
            xticklabel style = {font=\tiny},
            xlabel={ $n_{rel}$  (CSQA) },
           legend style={draw =none, legend columns=-1, font=\small, 
           nodes={scale=0.5, transform shape}
            }
        ]
        \addplot[orange,sharp plot] coordinates {(17,73.78)(7,73.78)(1,73.78) (0,73.78)};
        \addplot+ [
            color=blue!40!white,
            fill=blue!40!white,
            error bars/.cd,
            y dir=both,y explicit,
             error bar style={ 
                color=blue!40!black,
                fill=blue!40!black,},
            ] table[x=NR,y=RN-mean,y error=RN-variance]{\CNCSQAIHtest};
        \legend{ALBERT,+RN};
    \end{axis}
    \end{tikzpicture}
    \begin{tikzpicture}[baseline]
    \begin{axis}[
            name = plot1,
            ybar,
            bar width=.3cm,
            width=0.5*\columnwidth,
            height=3.5cm,
            enlargelimits=0.2, 
            ylabel style = {font=\tiny },
            xlabel style = {font=\scriptsize},
            xtick pos = left, 
            ytick pos = left, 
            symbolic x coords={17,7,1,0},
            xtick=data,
            y tick label style={
                        /pgf/number format/set thousands separator={}%
                        },
            ymin=62,
            ymax=66.7,
             yticklabel style = {font=\tiny},
             xticklabel style = {font=\tiny},
            xlabel={ $n_{rel}$ (OBQA)},
        ]
         \addplot[orange,sharp plot] coordinates {(17,63.47)(7,63.47)(1,63.47) (0,63.47)};
        \addplot+[
            color=blue!40!white, 
            fill=blue!40!white, 
            error bars/.cd,
            y dir=both,y explicit,
            error bar style={ 
                color=blue!50!black,
                fill=blue!50!black,},
            ] table[x=NR,y=RN-mean, y error=RN-variance]{\DataCNOBQAtest};
    \end{axis}
    \end{tikzpicture}
\end{adjustbox}
\caption{\label{fig:relation_diversity_csqa_ihtest} \normalfont Test set accuracy under different numbers of relation labels for \texttt{ConceptNet} on CSQA (left) and OBQA (right).
}
\end{figure} 

The competitive performance of \sw may be surprising, particularly with the relation-aware KG encoder RN, because \cn has access to a rich typed relation inventory while \swow does not. We investigate the impact of labelled relation types on CQA for RN, by ablating the number of relation types accessible to \cn: we grouped the 17 \cn relation types used in the models above into (a) seven coarse types (plus reverse relations) using the relational ontology of \newcite{liu_2004_conceptnet};\footnote{See Appendix~\ref{sec:relation_types_list} for details of 17 and 7 relation types.} (b) a single generic relation type (plus reverse relation). This version still contains one-hop and two-hop paths as well as reverse relations; and (c) removing all relation information from the model. 
Figure~\ref{fig:relation_diversity_csqa_ihtest} shows that merging \cn relations into broader types improves downstream task performance across data sets, with the best results achieved with one or seven relation types. Our results suggest that augmenting \swow with a rich label inventory may not be necessary for it to be used as a common-sense resource in downstream common-sense reasoning models. Nevertheless, labelling \swow with cognitively valid relation (or commonsense type) information in order to better understand the types of spontaneous associations humans express is an exciting avenue for future work.

\section{Conclusions}

We presented an in-depth analysis of the general and commonsense knowledge encoded in human word association norms (\swow), versus a traditional curated commonsense knowledge graph (\cn). We showed that the two knowledge resources differ systematically in their structure and content. We also showed that \sw brings comparable gains to \cn when applied to three commonsense question answering benchmarks, which is important as word associations are simpler than structured relations, and accordingly can be created more cheaply via crowd-sourcing and without the need for experts. Finally, we showed that \sw encodes situational commonsense knowledge as encoded in the human-created MCScript2.0 narratives  more directly than \cn; and that both KGs impose meaningful additional relations between concepts that were left implicit in the descriptions. There are several directions for future work, most notably extending our framework for characterizing  commonsense knowledge in the two KGs, exploring \sw relation types, developing means of consolidating the two knowledge graphs, and exploring downstream utility of KGs in less well-resourced languages than English, where the portability of the word association annotation methodology confers a substantial advantage.
\section*{Acknowledgements}

This work was supported in part by China Scholarship Council (CSC). We thank the reviewers for their valuable comments, and Simon De Deyne for insightful discussions. 

\bibliography{anthology}
\bibliographystyle{acl_natbib}

\clearpage
\appendix
\section{Relation types list}
\label{sec:relation_types_list}
For \texttt{ConceptNet}, we start with grouping the 31 original \cn relation types into 17 clusters following \citep{Wang2020ConnectingTD} (Table \ref{tab:17_cn_rel_types} left).  
We further group the 31 relation types into 7 by following \citep{liu_2004_conceptnet}, which grouping relation types into 7 categories, including \{things, spatial, events, causal, affective, functional, agents\} (Table \ref{tab:17_cn_rel_types} right).

\begin{table*}
    \footnotesize
    \begin{subfigure}[T]{0.6\textwidth}
    \begin{tabular}{ cp{3cm} | cp{3cm} }
    \toprule
    \multicolumn{4}{c}{\bf 17 relation types} \\
	\midrule
	1 &atlocation, locatednear & 10& usedfor \\\hline
	2 &capableof & 11 &receivesaction \\\hline
	3 &createdby & 12& madeof \\\hline
	4 &desires &  13 &partof, hasa\\\hline
	5 &hascontext & 14 &notdesires \\\hline
	6 &hasproperty & 15&notcapableof \\\hline
	7 &antonym, distinctfrom & 16 &isa, instanceof, definedas \\\hline
	8& relatedto, similarto, synonym& 17& causes, causesdesire,  motivatedbygoal \\\hline
	9& hassubevent, hasfirstsubevent, haslastsubevent,  hasprerequisite, entails, mannerof\\
	\bottomrule
    \end{tabular}
    \end{subfigure}%
    \begin{subfigure}[T]{0.4\textwidth}
    \begin{tabular}{cp{5cm}}
    \toprule
    \multicolumn{2}{c}{\bf 7 relation types}  \\\midrule
       1& capableof\\\hline
       2& usedfor, receivesaction \\\hline
    3&    atlocation, locatednear, hascontext, similarto \\\hline
      4&  causes, causesdesire, motivatedbygoal, desires \\\hline
    5&    antonym, distinctfrom, notcapableof, notdesires \\\hline
     6&   isa, hasproperty, madeof, partof, definedas, instanceof, hasa, createdby, relatedto, synonym \\\hline
      7&  hassubevent, hasfirstsubevent, haslastsubevent, hasprerequisite, entails, mannerof \\\bottomrule
    \end{tabular}
    \end{subfigure}
    \caption{Conflation of relation types in \cn to either 17 (left) or 7 (right) coarser grained groups.}
    \label{tab:17_cn_rel_types}
\end{table*}

\section{Negation List}
We use a list of negation markers to identify negated words and phrases. 
\label{sec:negation_list}
 The list of negation markers is:  (`no', `not', `none', `nor', `no one', `nobody', `nothing', `neither', `nowhere', `never', `hardly', `barely', `scarcely', `non', `without', `fail', `cannot', `cant', `no longer', `dont', `wont').
 Some negated triples sampled from \cn ans \swow are presented in Table~\ref{tab:negated_triples_cn_sw}.

\section{Hyper-parameters}
\label{sec:hyper_parameters}
We use the cross-entropy as the loss function with RAdam \citep{Liu2020ORadam} as optimizer to train all the models. We use as GELU \citep{hendrycks2016gelu} as the activation function. We report most important hyper-parameters in Table~\ref{tab:overall_hyperparameters}.

\section{Re-Production Results}
\label{sec:re_production_results}
For all KG-augmented models, we use the implementation from previous work \citep{Wang2020ConnectingTD}. \footnote{\url{https://github.com/INK-USC/MHGRN}} Table~\ref{tab:reimplementation_roberta_csqa_obqa} compares our re-run results to the original numbers reported in the respective papers. All our reproduced scores are comparable to or better than reported numbers. All of our experiments are run on single GPU of NVIDIA V100 16G.

\begin{table*}[!ht]
\footnotesize
\centering
\begin{tabular}{ll}
	\toprule
	\cn negated triples & \swow negated triples \\
	\midrule
	(antonym, still with us, no longer with us) &	(forwardassociated, love, non tangible) \\
    (antonym, awesome, fail) &	(forwardassociated, slight, barely) \\
    (antonym, able, cannot) &	(forwardassociated, real, not fake) \\
    (relatedto, nobody, no one) &	(forwardassociated, tedious, not fun) \\
    (synonym, zero, nothing) &	(forwardassociated, everything, nothing) \\
    (antonym, both, neither) &	(forwardassociated, with, without) \\
    (capableof, clues, lead nowhere) &	(forwardassociated, broke, no money) \\
    (causes, going to sleep, never waking up) &	(forwardassociated, lemon, not lime) \\
    (distinctfrom, hardly, easily) &	(forwardassociated, later, not now) \\
    (hassubevent, eat quickly, barely chew) &	(forwardassociated, punishment, not good) \\
    (derivedfrom, scarcely, scarce) &	(forwardassociated, none, no more) \\
    (causes, dying, non existence) &	(forwardassociated, succeed, fail) \\
    (hassubevent, eat healthily, dont eat junk food) &	(forwardassociated, unable, cannot) \\
    (relatedto, dare, you wont) &	(forwardassociated, obsolete, no longer exists) \\
	 \bottomrule
    \end{tabular}
    \caption{\label{tab:negated_triples_cn_sw} Examples of negated triples from \cn and \swow.}
\end{table*}

\begin{table*}[t!]
    \footnotesize
    \centering
    \begin{tabular}{ccc}
    \toprule
     Type &  Hyperparameter & Value \\ \hline
      General & \\  
              & batch size &  32/16/16\\ 
              & dropout & 0.1/0.2/0.1 \\
              & early stopping patience & 2 epochs \\
              & max sentence length & 80/84/300 \\
              & weight decay & 0.01 \\ \hline
     ALBERT-xxlarge-v2 & \\ 
         & learning rate & 1e-05 \\ \hline
     GconAttn &\\ 
         & learning rate & 3e-04/3e-04/1e-03 \\ 
         & MLP layers & 2 \\
         & hidden units & \{256, 128\} \\ 
         & concept embedding dimension & 1024 \\  \hline 
      RN & \\ 
         & learning rate & 1e-03/3e-04/1e-03 \\ 
         & MLP layers & 3 \\ 
         & hidden units & \{256, 256, 128\} \\ 
         & concept embedding dimension & 1024 \\ 
         & relation embedding dimension  &  100  \\\bottomrule
    \end{tabular}
    \caption{\label{tab:overall_hyperparameters} Hyperparameters for various models and data sets. Values split by ``/" follow the order of CSQA/OBQA/MCScript2.0.}
\end{table*}
\begin{table*}[!t]
 \centering
  \normalsize
    \begin{tabular}{lcccc}
	\toprule
	 Model & \multicolumn{2}{c}{CSQA} & \multicolumn{2}{c}{ OBQA}  \\
	 \cmidrule(lr){2-3} \cmidrule(lr){4-5}
	& \cite{Wang2020ConnectingTD} & Our re-run & \cite{Wang2020ConnectingTD} & Our re-run \\\hline 
	w/o KG & 68.69 (\(\pm \)0.56) & 70.46 (\(\pm \)0.18) & 64.80 (\(\pm \)2.37) & 64.47 (\(\pm \)3.01) \\
	+ GconAttn & 69.88 (\(\pm \)0.47) & 70.59 (\(\pm \)0.66) & 64.75 (\(\pm \)1.48) & 69.00 (\(\pm \)1.41) \\
	+ RN & 69.59 (\(\pm \)3.80) & 72.79 (\(\pm \)0.63) & 65.20 (\(\pm \)1.18) & 65.30 (\(\pm \)0.99) \\
	\bottomrule
    \end{tabular}
    \caption{Comparisons of our re-production of various KG-augmented models with previous work. The RoberTa-large encoder is used. We use the same provided code by \citep{Wang2020ConnectingTD} for CSQA and OBQA. Note that text representation on OBQA is the the average pooling over the hidden states of the last layer of RoberTa rather than `CLS' token representation. }
    \label{tab:reimplementation_roberta_csqa_obqa}
\end{table*}

\section{Full list of shortest paths }
\label{sec:full_list_of_shortest_paths}
Table~\ref{tab:shortest_path_full_list_cn_sw_183} presents all directed connected paths shown in Table~\ref{fig:narrative_kg_graphs_183} and their corresponding shortest paths retrieved \cn and \sw.

\begin{table*}[!t]
    \centering
    \footnotesize
    \begin{tabular}{llll}
	\toprule
	 & MCScript & \cn & \sw \\
	\midrule
	1 & (purchase, seed) & (purchase, sale, full, seed) & (purchase, need, seed) \\
	2 & (purchase, fertilizer) & (purchase, chain, garage, fertilizer) & (purchase, product, fertilizer) \\
	3 & (hole, cover) & (hole, opening, cover) & (hole, band, cover) \\
	4 & (flower, appear) & (flower, visit, appear) & (flower, become, appear) \\
	5 & (flower, pollinate) & (flower, pollen, pollinate) & (flower, bee, pollinate) \\
	6 & (flower, start) & (flower, open, start) & (flower, green, start) \\
	7 & (flower, bee) & (flower, bee) & (flower, bee) \\
	8 & (grow, continue) & (grow, carry, continue) & (grow, extend, continue) \\
	9 & (grow, vegetable) & (grow, fruit, vegetable) & (grow, vegetable) \\
	10 & (grow, plant) & (grow, plant) & (grow, plant) \\
	11 & (grow, weed) & (grow, field, weed) & (grow, weed) \\
	12 & (continue, pick) & (continue, carry, pick) & (continue, stick, pick) \\
	13 & (pollinate, bee) & (pollinate, pollen, bee) & (pollinate, bee) \\
	14 & (ripen, begin) & (ripen, change, action, begin) & (ripen, blossom, begin) \\
	15 & (garden, water) & (garden, earth, water) & (garden, water) \\
	16 & (garden, remove) & (garden, cricket, remove) & (garden, wart, remove) \\
	17 & (weed, remove) & (weed, remove) & (weed, remove) \\
	18 & (small hole, dig) & (small hole, mouse, hole, dig) & - \\
	19 & (bee, arrive) & (bee, branch, leave, arrive) & (bee, fly, arrive) \\
	20 & (day, make) & (day, clear, make) & (day, present, make) \\
	21 & (few seed, place) & - & - \\
	22 & (gardening tool, purchase) & (gardening tool, tool, lever, purchase) & - \\
	\bottomrule
\end{tabular}
    \caption{ Full list of paths from \cn and \sw for example shown in Figure~\ref{fig:narrative_kg_graphs_183}. - indicates path are not retrieved in the target KG.}.
    \label{tab:shortest_path_full_list_cn_sw_183}
\end{table*}


\end{document}